\pdfoutput=1
\documentclass[11pt]{article}

\usepackage[final]{acl}

\usepackage{times}
\usepackage{latexsym}

\usepackage[T1]{fontenc}
\usepackage[utf8]{inputenc}

\usepackage{microtype}

\usepackage{inconsolata}

\usepackage{graphicx}
\usepackage{booktabs}
\usepackage{amsmath}
\usepackage{algorithm}
\usepackage{algpseudocode}
\usepackage{amssymb}
\usepackage{tcolorbox}

\title{Mapping Smarter, Not Harder: A Test-Time Reinforcement Learning Agent That Improves Without Labels or Model Updates}

\author{
  \normalsize\textbf{Wen-Kwang Tsao}\thanks{\scriptsize Corresponding author.}, 
  \textbf{Yao-Ching Yu}\footnotemark[1], 
  \textbf{Chien-Ming Huang}
\\
\normalsize AI Lab, TrendMicro
\\
\normalsize{\texttt{\{spark\_tsao,yaoching\_yu\}@trendmicro.com}}
}

\begin{document}
\maketitle
\begin{abstract}
The Enterprise Intelligence Platform must integrate logs from numerous third-party vendors in order to perform various downstream tasks. However, vendor documentation is often unavailable at test time. It is either misplaced, mismatched, poorly formatted, or incomplete, which makes schema mapping challenging. We introduce a reinforcement learning agent that can self-improve without labeled examples or model weight updates. During inference, the agent first identifies ambiguous field-mapping attempts, then generates targeted web-search queries to gather external evidence, and finally applies a confidence-based reward to iteratively refine its mappings. To demonstrate this concept, we converted Microsoft Defender for Endpoint logs into a common schema. Our method increased mapping accuracy from 56.4\% (LLM-only) to 72.73\% (RAG) to 93.94\% over 100 iterations using GPT-4o. At the same time, it reduced the number of low-confidence mappings requiring expert review by 85\%. This new approach provides an evidence-driven, transparent method for solving future industry problems, paving the way for more robust, accountable, scalable, efficient, flexible, adaptable, and collaborative solutions.
\end{abstract}

\section{Introduction}

Enterprise IT environments are rapidly evolving toward proactive, agent-driven workflows. At the core of these systems lies a foundational requirement: the ability to efficiently process and semantically interpret logs from a wide range of third-party sources. Such capability enables agents to remain context-aware, access real-time data, and make well-informed decisions.

For cybersecurity use cases, enterprises must ingest massive volumes of logs—often terabytes per day—from heterogeneous sources such as firewalls, servers, endpoints, cloud applications, network flows, policy events, file operations, and API calls, in order to enable effective security operations and real-time threat detection~\cite{gartner2023}. The cost of Security Information and Event Management (SIEM) ingestion is substantial, exceeding \$500k annually for just 500~GB per day~\cite{techstrong2021,sentinel2024}. The challenge lies not only in scale, but also in achieving semantic consistency across dozens to hundreds of disparate event types. Failures in log correlation and schema normalization have contributed to catastrophic breaches—including Target (2013) and Equifax (2017)—where overlooked alerts and misaligned schemas resulted in damages exceeding \$1~billion~\cite{securitylogging2023}.

The emergence of large language models (LLMs) with robust natural language processing capabilities has the potential to transform third-party log integration. These models could dramatically reduce the need for labor-intensive processes that require experts. The full log integration pipeline comprises four stages: processing raw logs into structured data, mapping source schemas to target schemas, generating data transformation code, and deploying to production with ongoing monitoring. Schema mapping is the critical decision point that underpins the success of the entire integration process. This paper focuses on the practical challenge of schema mapping, a topic that is often overlooked in current literature.

Our focus is on a scenario in which enterprises ingest new data into their platforms. Therefore, the target schema is usually well-documented and stable. By contrast, incoming data source schemas are often poorly documented, typically due to their origin in legacy systems or outdated software versions. Unlike conventional machine learning tasks—where the challenge is extracting key features from abundant data—the difficulty here is the opposite: the source provides too little context. Because these vendor schemas lack labeled training data, fine-tuning or other supervised methods are impractical. Although expert review remains necessary, it is resource-intensive and must be carefully prioritized.

To address these challenges, we propose a test-time reinforcement learning (RL) agent that can self-improve without updating model weights or relying on pre-defined labeled data. Our approach is inspired by the TTRL framework \cite{zuo2025ttrl}, which improves model performance on unlabeled data through test-stage rewards. In the absence of ground-truth labels, we introduce a confidence-based score as a proxy reward signal to guide the agent toward more accurate mappings. Our design explicitly targets industrially practical constraints: model updates are costly, GPU-intensive, and operationally complex. Instead, our method adapts the system prompt at inference time by iteratively enriching the context in a verbal RL-driven manner, thereby enabling continual self-improvement under real-world deployment conditions.

The confidence-based proxy reward is both intuitive and interpretable. The agent identifies conflicts and ambiguities in its prior mapping attempts and formulates targeted search queries to gather external evidence. Confidence scores then serve as reward signals to determine whether the collected evidence should be retained or discarded, guiding iterative refinement. Furthermore, the agent produces transparent reasoning traces, enabling human experts to focus their review on low-confidence mappings, thereby reducing manual verification costs while preserving overall reliability.

This method has several advantages over existing approaches. First, it achieves significantly higher mapping accuracy through iterative refinement, reducing reliance on static, one-time attempts. Second, it uses confidence-based rewards instead of ground truth labels, enabling effective learning in settings where labeled data is unavailable. Finally, it promotes transparency by revealing its reasoning and evidence collection processes. This empowers security experts to understand decisions and prioritize low-confidence mappings.

Overall, our key contributions are: 1) identifying the challenge of schema mapping where prompting, fine-tuning, and retrieval-augmented generation all face limitations and no ground truth labels are available, and 2) proposing a test-time reinforcement learning framework that enables an agent to improve accuracy over time. This approach opens a new research direction by using confidence as a proxy signal to guide the agent’s learning process.

\begin{figure}[!htb]
  \includegraphics[width=\columnwidth]{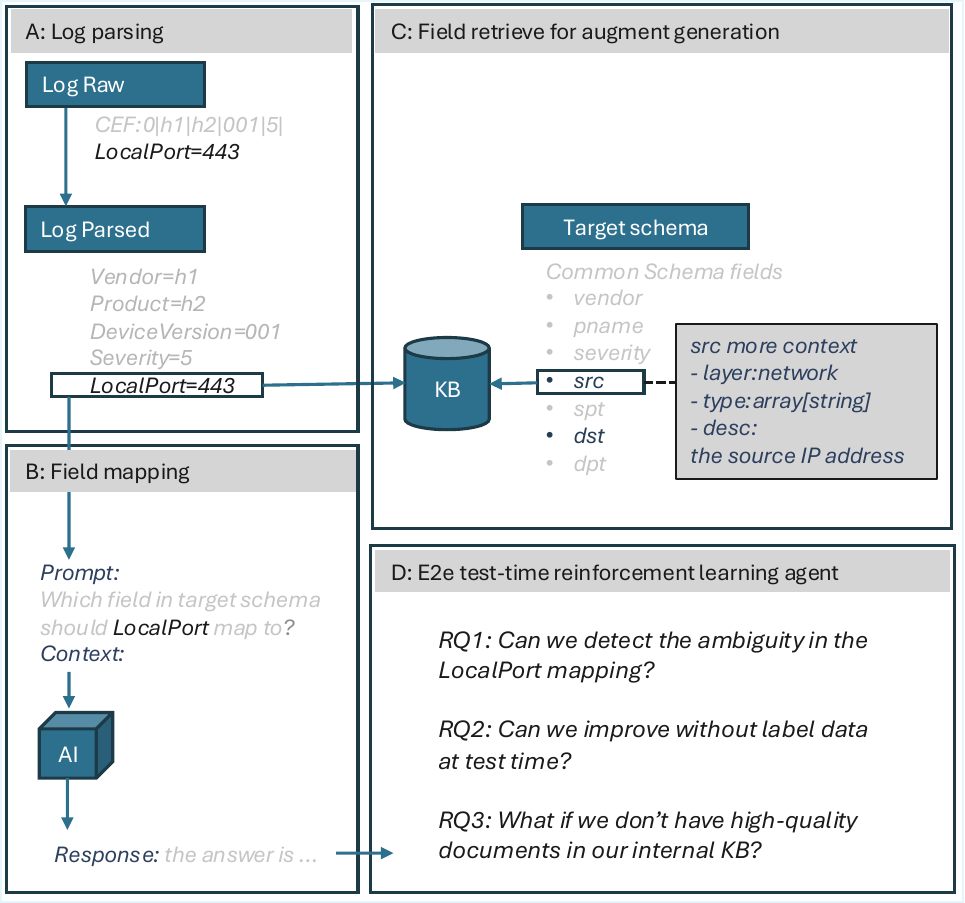}
  \caption{Position of our test-time RL agent relative to prior work in the schema matching pipeline. In Section A, Logparser-LLM \citet{zhong2024logparser} handles raw-to-structured parsing. In section B, Schema-Matching-LLM \citet{parciak2024schema} shows baseline one-shot mapping capability. In section C, ReMatch \citet{sheetrit2024rematch} adds retrieval when full documentation exists. MatchMaker \citet{seedat2024matchmaker} pre-embeds target schemas for reuse. In section D, Self-consistency \citet{wang2022self} improves chain-of-thought reasoning. Search-R1 \citet{jin2025search} enhances LLM reasoning with search-augmented RL. This example shows that the log LocalPort is ambiguous for decision making; we need more context to determine whether this field should map to src or dst. Our research uniquely extends rightward beyond traditional enterprise knowledge bases to handle newly seen logs with ill-formatted or incomplete documentation. Unlike fine-tuning approaches that require labeled data and risk overfitting, our agent operates without test-time labels, conducting internet searches to gather evidence outside the enterprise KB scope. This addresses the critical gap where traditional methods fail on unseen vendor schemas with insufficient documentation.}
  \label{fig:figrl_process}
\end{figure}

\section{Related Work}

The foundation of schema mapping begins with converting raw logs into structured data. \citet{zhong2024logparser} introduced LogParser-LLM, which addresses the initial step of our pipeline by transforming unstructured log messages into key-value pairs. Their hybrid approach combines an LLM-based template extractor with a prefix-tree clustering method, achieving efficient parsing; the authors report \textasciitilde{}272.5 LLM calls amortized across large log sets. This work eliminates the need for hyperparameter tuning or labeled data, enabling rapid adaptability to new log formats.

LogParser-LLM's contribution is complementary to our work: while they focus on the raw-to-structured parsing phase. In contrast, our approach begins where their process ends: once logs have already been parsed into structured records, we take these structured logs and further map them into standardized, common schemas that enable consistent downstream analysis, correlation, and integration across diverse sources. 

\citet{parciak2024schema} conducted a comprehensive experimental study of LLM capabilities in schema matching, focusing on off-the-shelf performance using only element names and descriptions. Their work provides crucial insights into the baseline capabilities of LLMs for schema matching tasks, demonstrating that context scope significantly affects performance—neither too little nor too much context leads to optimal results.

Their findings directly inform our approach in several ways: (1) they validate that LLMs can perform meaningful schema matching without requiring data instances, which aligns with our privacy-sensitive enterprise scenarios; (2) their context optimization insights guide our prompt engineering; and (3) their baseline performance metrics provide a foundation for measuring the improvements achieved by our reinforcement learning approach. However, their work focuses on one-shot matching capabilities, while our approach addresses the iterative improvement challenge through test-time learning.

ReMatch, introduced by \citet{sheetrit2024rematch}, leverages retrieval-augmented prompting to reduce the target schema search space before matching. Their approach assumes the availability of comprehensive documentation and uses embedding-based retrieval to balance recall and precision in schema matching tasks. ReMatch demonstrates strong performance in healthcare schema matching scenarios where complete documentation is available.

Our work differs from ReMatch in a fundamental assumption: while ReMatch operates in environments with well-documented schemas and comprehensive knowledge bases, our approach is designed for practical enterprise scenarios where such documentation is often incomplete or unavailable. ReMatch's retrieval mechanism works within a closed set of known mappings, whereas our agent dynamically discovers and accumulates evidence from external sources to handle previously unseen vendor schemas.

\citet{seedat2024matchmaker} introduced MatchMaker, which decomposes schema matching into candidate generation, refinement, and confidence scoring steps using LLMs and vector retrieval. Their approach focuses on embedding target schemas for better retrieval and future reuse, generating synthetic in-context examples on-the-fly for zero-shot self-improvement.

MatchMaker's compositional approach and confidence scoring align with our methodology, but their self-improvement mechanism differs significantly from ours. While MatchMaker generates synthetic examples for in-context learning, our approach accumulates real-world evidence through external search and interaction. MatchMaker's focus on target schema embedding optimization complements our source schema adaptation strategy, suggesting potential for hybrid approaches in future work.

Two key techniques underpin our reinforcement learning framework: search-enhanced reasoning and self-consistency validation. \citet{jin2025search} demonstrated how external search can enhance LLM reasoning capabilities, providing a foundation for our evidence collection mechanism. Their Search-R1 approach shows that targeted web search can significantly improve reasoning accuracy in complex tasks.

\citet{wang2022self} established that consistency across multiple reasoning paths serves as a reliable indicator of correctness, forming the basis of our confidence-based reward system. Their self-consistency approach demonstrates that taking the majority answer from multiple sampled reasoning paths significantly improves accuracy, which directly informs our confidence score calculation.

A closely related line of research is the \textit{Reflexion} framework~\cite{shinn2023reflexion}, which introduces verbal reinforcement learning as a way for language agents to improve through self-generated feedback rather than model weight updates. Reflexion agents reflect on task feedback, store these reflections in episodic memory, and leverage them to guide future decisions. This approach demonstrates that linguistic feedback—whether scalar or free-form—can significantly enhance agent performance across sequential decision-making, coding, and reasoning tasks, achieving state-of-the-art results on benchmarks such as HumanEval. Our work shares a similar philosophy in emphasizing memory- and feedback-driven improvement at test time; however, we focus specifically on schema mapping under industrial constraints where ground truth labels and comprehensive documentation are unavailable. In this setting, we extend the notion of verbal reinforcement learning by integrating external evidence collection with confidence-based rewards to refine mappings dynamically.

Our work addresses the critical gap where existing methods fail on newly encountered vendor schemas with ill-formatted or incomplete documentation. Unlike prior approaches that operate within controlled environments with well-documented schemas, our test-time reinforcement learning agent adapts dynamically by accumulating external evidence without requiring model updates or labeled training data.

\section{Methodology}

\subsection{Problem Formulation}

We formally define the schema mapping problem as follows. Given a source schema $S = \{s_1, s_2, \ldots, s_n\}$ and a target schema $T = \{t_1, t_2, \ldots, t_m\}$, we seek to find a mapping function $f: S \rightarrow T \cup \{\emptyset\}$ such that $f(s_i) = t_j$ represents a semantic correspondence between source field $s_i$ and target field $t_j$, or $f(s_i) = \emptyset$ indicates no corresponding field exists in the target schema.

Our objective is to maximize the correctness of this mapping function: $\max_{f} \sum_{i=1}^{n} \mathbb{I}[f(s_i) = f^*(s_i)]$ where $f^*(s_i)$ represents the ground truth mapping and $\mathbb{I}[\cdot]$ is the indicator function. However, in practical deployment scenarios—particularly when processing third-party vendor logs ground truth mappings $f^*$ are unavailable.

To address this challenge, we use confidence scores as a proxy for correctness. We define confidence $C(f(s_i))$ as the consistency of mapping predictions across multiple inferences, serving as a surrogate objective: $\max_{f} \sum_{i=1}^{n} C(f(s_i))$.

\subsection{Test-Time Reinforcement Learning Framework}

% Smaller gray comments with //
\algrenewcommand\algorithmiccomment[1]{\hfill{\footnotesize\textcolor{gray}{// #1}}}

\begin{algorithm*}[ht]
\caption{Iterative RL Schema Mapping with Confidence Improvement}
\label{alg:schema_mapping_rl}
\textbf{Inputs:} Source schema $S = \{s_1, \dots, s_n\}$; target schema $T = \{t_1, \dots, t_m\}$; iteration limit $\alpha$ (default: $100$); conflict detection attempts $n$ (default: $3$); initial evidence context $E = \emptyset$; generative LLM $\mathcal{F}$.

\textbf{Outputs:} Mapping function $f$; refined evidence context $E$.

\begin{algorithmic}[1]
\For{$i \gets 1$ to $\alpha$}
    \State $f \gets \mathcal{F}(S, E)$ \Comment{Generate initial mappings using current evidence}
    \State $C \gets$ ConflictDetection$(f, n)$ \Comment{Detect inconsistent mappings}
    \For{each source field $s_i \in C$}
        \State $Q_{s_i} \gets$ QueryFormulation$(s_i)$ \Comment{Formulate search queries}
        \State $e_{s_i} \gets$ EvidenceCollection$(Q_{s_i})$ \Comment{Collect external evidence}
        \State $r_{s_i} \gets$ ConfidenceEvaluation$(f(s_i), e_{s_i})$ \Comment{Evaluate new confidence}
        \State $r_{\text{prev}} \gets$ Confidence$(f(s_i))$ \Comment{Retrieve previous confidence}
        \If{$r_{s_i} > r_{\text{prev}}$}
            \State $E \gets$ ContextUpdate$(E, e_{s_i}, r_{s_i})$ \Comment{Update context if confidence improves}
        \EndIf
    \EndFor
\EndFor
\State \Return $f$, $E$
\end{algorithmic}
\end{algorithm*}

Our test-time reinforcement learning agent improves schema mapping policy's accuracy by iteratively collecting evidence and reinforcing useful context. The agent starts from an empty evidence set, identifies inconsistencies through multiple mapping attempts, and formulates targeted search queries for ambiguous fields. Collected evidence is added to the context, and confidence-based reward signals guide which evidence to retain or discard. The complete algorithm is detailed in Algorithm~\ref{alg:schema_mapping_rl}.

\begin{figure*}[ht]
\centering
\includegraphics[width=0.8\textwidth]{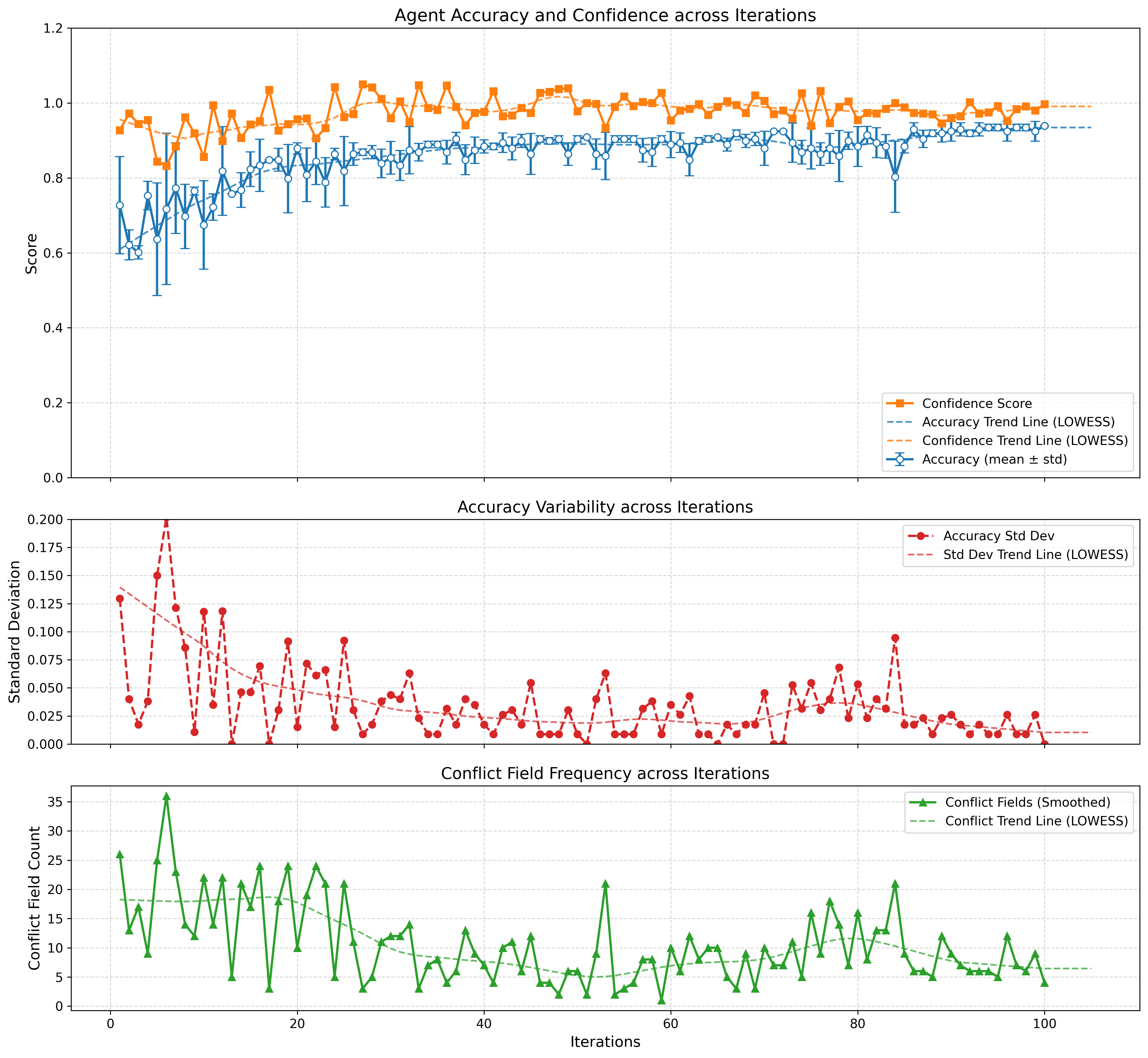}
\caption{Comprehensive performance analysis of the reinforcement learning agent over 100 iterations. Top: Accuracy improves from 72.73\% to 93.94\%, while confidence trends upward and approaches 1.0, highlighting a persistent overconfidence gap. Middle: Accuracy variability is high in early iterations but stabilizes over time. Bottom: Conflict field count decreases from 26 to 4(85\% reduction), demonstrating reduced ambiguity as evidence accumulates.}
\label{fig:accuracy}
\end{figure*}

\subsection{Reinforcement Learning Formulation}

To formalize our approach, we define the reinforcement learning components as follows:

\textbf{State}: The current state $s_t$ consists of the schema mapping hypothesis from source to target fields and the collected evidence at time $t$. Specifically: $s_t = \{M_t, E_t\}$ where $M_t$ is the current mapping hypothesis and $E_t$ is the set of evidence collected up to time $t$.

\textbf{Action}: The action $a_t$ involves selecting a field or conflict to investigate and executing a targeted search query. The action space includes all possible fields that could be investigated and the possible search queries that could be formulated.

\textbf{Reward}: The reward $r_t$ is derived from the change in confidence score after adding new evidence: $r_t = C_{t+1} - C_t$ where $C_t$ is the confidence score at time $t$. The confidence score serves as a proxy for accuracy when ground truth labels are unavailable.

\textbf{Policy}: The agent’s policy $\pi(a|s)$ specifies the suggested mapping from source schema fields to target schema fields. While the policy can be implemented in various ways, one practical approach is to leverage an LLM’s reasoning process. In this setup, the LLM performs the mapping by combining its prior knowledge with contextual evidence, enabling the system to detect conflicts and generate targeted search queries. We define a conflict as a disagreement among the $n$ prompt-variant predictions for the same source field within a single iteration (as opposed to cross-run drift).

\textbf{Learning}: The goal of learning is to adapt a policy’s behavior in a beneficial direction by leveraging rewards that reflect the situation. Instead of updating the LLM's model weights, learning takes place through the accumulation of useful evidence in the agent's context. Evidence that increases confidence is preserved, while unhelpful evidence is discarded. This process reflects a form of verbal, memory-based learning in which the agent's knowledge base is continuously refined.

We use verbal RL in the Reflexion sense—policy adaptation via context/memory rather than parameter updates; rewards are intrinsic confidence deltas, not ground-truth returns.

\section{Experiment}

\subsection{Setup}

We evaluated our approach using two schemas: the \textbf{source} schema was the \textit{Microsoft Defender for Endpoint} schema containing 195 fields, and the \textbf{target} was our \textit{Common Schema} with 137 fields. The \textbf{ground truth} consisted of 66 manually verified field mappings, curated collaboratively by domain, threat intelligence, and product experts to cover a diverse range of field types and complexity levels. We employed \textbf{GPT-4o} as the underlying model and assessed performance using two key metrics: (1) \textbf{Confidence score}---measuring the consistency of predictions across multiple attempts within a single iteration, with special handling for empty outputs; and (2) \textbf{Accuracy}---defined as the percentage of correctly predicted mappings among the 66 verified pairs.

For each field mapping iteration, the model was executed three times (n=3), and both the confidence score and accuracy were computed. While accuracy depends on the ground truth and is used solely for evaluation, confidence scores—which can be calculated without ground truth—are leveraged to guide the reinforcement learning process and are therefore suitable for production deployment. 

\textbf{Confidence Score Calculation:} We calculate confidence as the consistency of predictions across multiple attempts using a modified frequency-based approach. For a given field, if we collect predictions $P = [p_1, p_2, p_3]$ across three iterations, the confidence score is: $C = \frac{\text{count}(\text{most\_frequent\_prediction})}{\text{adjusted\_total}}$ where the adjusted total accounts for empty predictions with reduced weight (0.5 instead of 1.0). The design aligns with the findings of \citet{kalai2025language}, who argue that current training and evaluation paradigms implicitly reward guessing behavior in language models, leading to hallucinations. By contrast, our scoring mechanism provides a quantitative incentive for uncertainty acknowledgment, steering the model toward more trustworthy behavior.

We prepare two baselines for comparison.  
\textbf{Baseline 1:} We used GPT-4o with a single-shot prompt containing only the field name and value, resulting in an accuracy of 56.36\%.  
\textbf{Baseline 2:} We used GPT-4o with a single-shot prompt enhanced with additional field descriptions, data types, and sample data context from our internal knowledge base, resulting in an accuracy of 72.73\%.

\subsection{Performance Improvements}

Starting from a \textbf{Baseline 2:} accuracy of 72.73\% with GPT-4o in the first iteration, our method achieved significant improvements throughout the experiment. By the end of the 100-iteration experiment, the model demonstrated consistent performance with accuracy reaching 93.94\%.

Our analysis of the 100-iteration experiment reveals several key performance characteristics. GPT-4o collects 81 evidence tuples across 100 iterations, with the most significant accuracy gains occurring in the early stages—for example, iteration 1 shows a +21.21\% gain, iteration 5 achieves +9.60\%, and iteration 6 yields +10.71\%. The system makes decisions solely based on confidence scores, without referencing accuracy at any point during execution. 

This reflects real-world conditions in which ground truth labels are unavailable. In these scenarios, accuracy can only be measured retrospectively to confirm whether the solution performed correctly. Notably, in 19 of the 100 iterations, the system rejected newly proposed evidence, demonstrating that it learned to recognize when gathering additional information was unlikely to improve results. This selective behavior preserves a high-quality evidence set, which is critical not only for validating outcomes but also for providing transparency into how the agent arrived at its decisions.

By the conclusion of the 100 iterations, the agent resolved 22 conflicts, only 4 fields were flagged as low-confidence and requiring expert review—down from 26 initially—representing an 85\% reduction in the verification burden on security analysts and allowing them to focus on the most ambiguous or critical cases.

To evaluate the robustness of our approach, we ran the full 100-iteration process multiple times. The final accuracy consistently reached 93-94\%, with minimal variance (standard deviation less than 0.01) in the later iterations. This demonstrates that the improvement is reliable and not due to chance.

\begin{figure}[tbp]
\centering
\includegraphics[width=\linewidth]{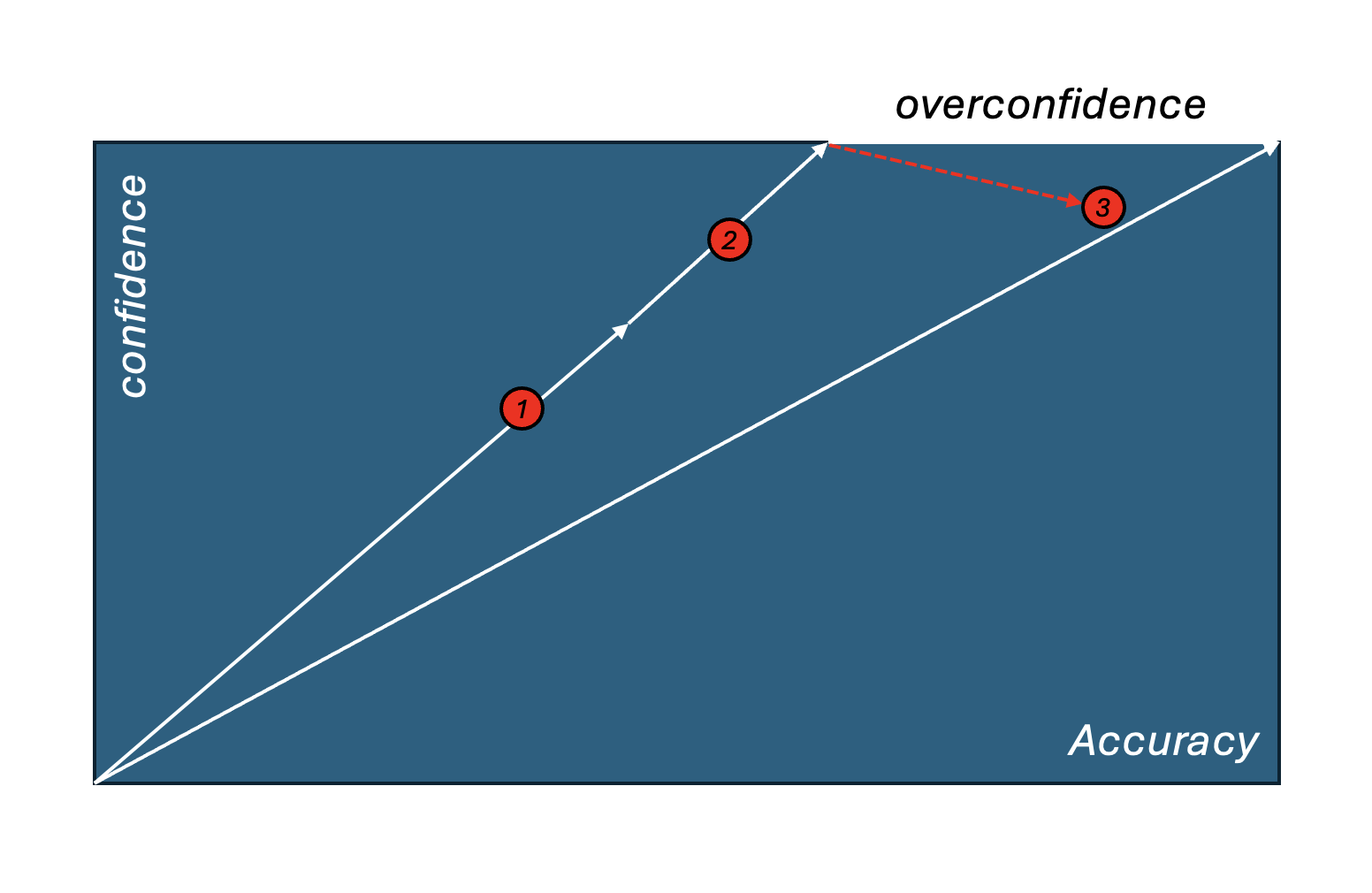}
\caption{
The relationship between confidence and accuracy. Path 1 shows the starting point. Path 2 illustrates how confidence as a proxy reward improves accuracy. Path 3 highlights the challenge of overconfidence, where confidence saturates while accuracy lags behind. Therefore, more engineering and research efforts are needed to bring the confidence curve down. The diagonal line indicates perfect calibration, and the curves above it show overconfidence.
}
\label{fig:confidence-curve}
\end{figure}

\subsection{Evidence Collection Analysis}

For this experiment, we equipped the AI agent with a tool that enables internet searches for additional facts and evidence. Specifically, we used a generic search utility powered by Bing. In each iteration, the agent analyzes conflicts identified across three prompt variants, formulates a targeted search query, and retrieves relevant results from the tool. This setup demonstrates one possible method of evidence collection; other approaches could include consulting human experts, performing advanced reasoning, or validating findings against production data.

We organized the collected evidence into three-element tuples, each comprising the detected conflict, the resolution plan, and the retrieved evidence. This structured representation ensures systematic evidence collection and evaluation. Evidence is retained only if it demonstrably increases the LLM’s confidence in its mapping decisions. Helpful evidence contributes in 4 ways, (1) by enhancing awareness through the identification of ambiguous fields and conflicting mappings; (2) by enabling self-correction through improved reasoning about relationships among fields; (3) by improving clarity via authoritative information on how fields are defined and used in both the common schema and vendor-specific schemas (e.g., Microsoft Defender); and (4) by revealing context-dependent mappings, in which the correct correspondence varies by scenario and requires additional contextual understanding.

\section{Conclusion}

This paper presents a test-time learning framework in which a reinforcement learning agent improves schema mapping accuracy through iterative evidence collection and context refinement—without requiring model weight updates or pre-collected training data. Unlike conventional approaches that rely on all labeled data before inference, our method actively gathers new evidence during test-time to refine its decisions. This design eliminates the high computational cost, GPU dependency, and potential instability associated with model retraining, making it more practical for real-world deployment. A confidence-based score serves as a proxy reward for accuracy, providing a novel mechanism that evaluates model consistency across multiple mapping attempts within a single iteration. Through systematic evidence collection and evaluation, the agent resolves mapping ambiguities with transparent reasoning that facilitates expert review and validation. Applied to Microsoft Defender for Endpoint logs mapped to a common schema, our approach improves accuracy from 72.73\% to 93.94\%, while reducing the number of fields requiring expert review by 85\%—from 26 to 4.

A key insight from our work is the use of confidence scores as a proxy for ground-truth labels, as conceptually illustrated in Figure~\ref{fig:confidence-curve}. While confidence can effectively guide accuracy improvement, our results reveal persistent overconfidence, where confidence values exceed actual accuracy. This observation underscores the need for better confidence definition and calibration. In our current approach, confidence is measured as the consistency of predictions across three trials. Future research could extend this by (1) employing multiple models to generate diverse predictions and compute ensemble-based confidence, or (2) directly prompting LLMs to provide explicit self-assessed confidence scores alongside their outputs. Additional experiments extending the number of inference attempts from three to ten show encouraging signs of reducing the overconfidence gap (see Appendix~B), albeit with increased computational cost. These enhancements could yield better-calibrated confidence measures that more accurately reflect true prediction quality.

\section*{Limitations}

While our approach demonstrates significant improvements, several limitations should be acknowledged:

\begin{enumerate}
\item \textbf{Evidence Quality Dependency}: The system's performance is dependent on the quality and availability of external evidence sources. In domains where documentation is sparse or inconsistent, the improvement may be limited.

\item \textbf{Computational Overhead}: The iterative evidence collection process requires multiple LLM calls and external searches, which may increase computational costs compared to single-shot approaches.

\item \textbf{Domain Specificity}: Our evaluation focuses on cybersecurity schemas. While the approach is general, validation in other domains (healthcare, finance, etc.) would strengthen the generalizability claims.

\item \textbf{Scope Limitations}: We focus on 1-to-1 mappings to address the core challenge of knowledge scarcity. Extension to more complex mapping cardinalities (1-to-N, N-to-M) remains future work.
\end{enumerate}

\section*{Ethical Considerations}
This study does not involve human subjects or personal data. However, the method collects web evidence at test time, which could occasionally include malicious or adversarial content. Practitioners reproducing this system should apply input sanitization, source filtering, and sandboxing to prevent prompt-injection or security risks.

\bibliography{custom}

\newpage

\appendix

\section{Case Study: Direction-Sensitive Port Mapping (Iteration 49)}

To illustrate the practical difficulty of schema mapping, we examine a representative example from iteration~49 involving the common schema field \texttt{dpt} (destination port). The field \texttt{dpt} is defined in the Trend Micro Common Schema as ``the service destination port of the private application server (dstport).'' In Microsoft Defender for Endpoint, however, two candidate fields exist: \texttt{LocalPort} (TCP port on the local device used during communication) and \texttt{RemotePort} (TCP port on the remote device being connected to).

\paragraph{Mapping ambiguity.}
Both candidates are semantically plausible depending on the traffic direction. For \emph{outbound} connections, the local device is the source, so the destination port resides on the remote endpoint—corresponding to \texttt{RemotePort}. Conversely, for \emph{inbound} connections (e.g., an RDP session initiated by a remote host), the local device becomes the destination, and thus the correct mapping is \texttt{LocalPort}. Without an explicit indicator of connection direction, an LLM can easily misassign the field.

\paragraph{Why fine-tuning does not help.}
Simply fine-tuning model weights cannot reliably resolve this ambiguity. The correct mapping is not a memorized association (\texttt{dpt} $\rightarrow$ \texttt{RemotePort}) but a conditional rule that depends on runtime context—information that is absent from the training corpus. Updating parameters may reinforce spurious correlations rather than teach the model to reason over directionality, resulting in brittle behavior across different network scenarios.

\paragraph{Evidence-based reasoning.}
During iteration~49, the agent retrieved definitional evidence clarifying that:
\begin{itemize}
  \item \texttt{LocalPort} refers to the port on the local device.
  \item \texttt{RemotePort} refers to the port on the remote device being connected to.
  \item \texttt{dpt} represents the service destination port.
\end{itemize}
Although this evidence did not directly yield the final mapping, it revealed that the correct correspondence varies by communication context and motivated the agent to infer direction from auxiliary fields such as \texttt{RemoteIP}, \texttt{LocalIP}, and \texttt{ActionType}. The agent’s confidence increased from 0.67 to 1.0, reflecting a clearer and more consistent conceptual understanding.

\paragraph{Categorization of helpful evidence.}
According to the taxonomy defined in the main paper, this instance exemplifies \textbf{Category~(4): revealing context-dependent mappings}. The evidence was helpful because it exposed the conditional nature of the mapping—showing that correctness depends on dynamic context rather than static field alignment—and thereby guided subsequent reasoning toward more generalizable, context-aware mapping rules.

\paragraph{Derived practical rule.}
\[
\texttt{dpt} =
\begin{aligned}
&\texttt{RemotePort}, && \text{if Direction = Outbound;}\\
&\texttt{LocalPort},  && \text{if Direction = Inbound;}\\
&\text{(defer/flag)}, && \text{if Direction is unknown.}
\end{aligned}
\]
This case highlights that progress in schema alignment arises not from parameter updates but from the integration of structured evidence and contextual inference.

\section{Overconfidence Mitigation Experiments}

To address the overconfidence phenomenon observed in our main experiments (Figure~\ref{fig:confidence-curve}), we conducted an additional study to examine how increasing the number of inference attempts affects confidence calibration. Our hypothesis was that a larger number of inference samples would improve statistical robustness and reduce systematic overconfidence.

\subsection{Experimental Setup}

We extended the original setup by increasing the number of inference attempts per iteration from 3 to 10. This change allows for more reliable confidence estimation through better sampling of the model’s output distribution. All other experimental parameters remained identical to the main study: the same source and target schemas (Microsoft Defender for Endpoint with 195 fields and the Common Schema with 137 fields), the same 66 manually verified ground-truth mappings, and the same GPT-4o model configuration.

The confidence score computation followed the same principle of prediction consistency but with a larger sample size (10 vs.\ 3). This design provides a more statistically grounded measure of uncertainty and is expected to yield better-calibrated confidence estimates.

\subsection{Results and Analysis}

Table~\ref{tab:overconfidence_comparison} presents a comparison between the 3- and 10-inference settings.

\begin{table*}[t]
\centering
\begin{tabular}{lccc}
\toprule
\textbf{Setting} & \textbf{Final Accuracy} & \textbf{Mean Confidence} & \textbf{Low-Confidence Fields} \\
\midrule
3 Inferences & 93.94\% & 95.2\% & 26 $\rightarrow$ 4 (-85\%) \\
10 Inferences & 92.1\% & 89.3\% & 31 $\rightarrow$ 6 (-80\%) \\
\bottomrule
\end{tabular}
\caption{Comparison of overconfidence mitigation between 3 and 10 inferences per iteration. The 10-inference setting improves calibration by bringing confidence values more closely aligned with actual accuracy.}
\label{tab:overconfidence_comparison}
\end{table*}

Increasing the number of inference attempts from 3 to 10 effectively narrows the gap between confidence and accuracy. With 10 inferences, the mean confidence (89.3\%) aligns much more closely with the observed accuracy (92.1\%), whereas with 3 inferences, confidence (95.2\%) noticeably exceeds accuracy (93.94\%), indicating mild overconfidence. 

This calibration improvement comes with a modest cost: while the confidence estimates become more realistic, overall accuracy decreases slightly (93.94\% → 92.1\%). Nevertheless, the more conservative confidence levels offer better guidance for expert review prioritization. The reduction in low-confidence fields remains substantial under both settings (85\% and 80\% reductions), demonstrating that the key benefit—reducing expert review effort—is preserved.

\subsection{Implications and Trade-offs}

These findings highlight a clear balance between calibration quality and computational cost. Increasing the number of inference attempts leads to better-calibrated confidence scores but requires more computation. In practical terms:
\begin{itemize}
    \item \textbf{For high-precision scenarios}, increasing attempts (e.g., to 10) provides more reliable uncertainty estimates and tighter alignment between confidence and accuracy.
    \item \textbf{For cost-sensitive deployments}, even three attempts already achieve strong accuracy and substantial reduction in expert workload, making it a highly efficient operational setting.
\end{itemize}

These results demonstrate improved calibration—confidence more accurately tracks empirical accuracy as the number of inferences increases—without the need for additional metric reporting.

\section{Prompt Architecture}

Table~\ref{tab:prompt_overview} summarizes the three prompts used in our schema mapping system. Each serves a distinct function in guiding the agent’s reasoning process across sessions and mapping iterations.

\begin{table*}[t]
\centering
\begin{tabular}{llp{6cm}p{3.2cm}}
\toprule
\textbf{\#} & \textbf{Prompt Name} & \textbf{Purpose} & \textbf{Usage} \\
\midrule
1 & System Prompt & Defines the AI agent's role as a cybersecurity data expert specializing in schema mapping. Establishes reasoning rules and guidelines. & Loaded once per session \\
2 & User Prompt & Carries the per-request payload: (a) curated facts from prior conflicts (if any), (b) RAG-assembled source/target schema context (descriptions, sample values, types), and (c) the mapping task/question. Enforces a strict XML response (CSV decision, 1–5 confidence, reasoning) for deterministic parsing. & For every mapping request \\
3 & Search Prompt & Generates targeted Internet search queries string to resolve ambiguous field mappings using prior conflict information. & Invoked only during conflict resolution \\
\bottomrule
\end{tabular}
\caption{Overview of the three prompts used in the schema mapping system, detailing their roles and usage.}
\label{tab:prompt_overview}
\end{table*}

\vspace{0.5em}  % adds a little space before the next subsection

\noindent\textbf{System Prompt Example.}
The system prompt defines the AI agent’s expertise and reasoning framework. It is shown below for reference.

\vspace{0.3em}  % adds breathing space before the box

\begin{tcolorbox}
\small
You are a Trend Micro cybersecurity data expert specializing in Trend Micro’s Common Schema across multiple layers, including endpoint, network, messaging, and cloud.  
You have extensive expertise in processing and mapping third-party product and log schemas to Trend Micro’s Common Data Schema, enabling cross-log correlation, advanced threat detection, and compliance reporting.

You routinely perform professional schema mapping for new third-party log sources with a focus on accuracy. Your approach follows a layered reasoning process:  
(1) identify core entities such as IP addresses, filenames, and hashes (e.g., SHA1);  
(2) narrow down candidate fields based on data flow direction and context;  
(3) make precise mapping decisions supported by semantic consistency.  

{For example:}  
-- \texttt{src\_ip} vs.\ \texttt{dst\_ip} depends on whether traffic is inbound or outbound.  
-- \texttt{SHA1} hashes may represent \texttt{parent\_process\_sha1}, \texttt{launched\_process\_sha1}, or \texttt{dropped\_object\_sha1}.  

If no suitable mapping exists, respond professionally with \texttt{NOT\_COVERED}.
\end{tcolorbox}

\end{document}